\documentclass{tlp}
\usepackage{amsmath,amssymb,amsfonts}%
\usepackage[title]{appendix}%
\usepackage{xcolor}%
\usepackage{textcomp}%
\usepackage{stmaryrd}
\usepackage{enumitem}
\usepackage{mathtools}
\usepackage[all]{xy}
\usepackage{orcidlink}
\newcommand{\HT}{\rm HT}
\def\Lang{\mathcal{L}}
\def\at{\ensuremath{{\cal AT}}}

\newcommand{\den}[1]{\llbracket \, #1 \, \rrbracket}
\newcommand{\fden}[1]{\langle\hspace{-2pt}\langle \, #1 \, \rangle\hspace{-2pt}\rangle}
\newcommand{\SM}[1]{\text{\rm \em SM}(#1)}
\newcommand{\SMR}[2]{\text{\rm \em SM}_{#2}(#1)}

\newcommand{\inh}{\mathbin{\hat{\in}}}
\newcommand{\ideal}[1]{\,\downarrow\!\!#1}
\newcommand{\idealset}[1]{{}^\downarrow\!#1}
\def\cH{\mathcal{H}}
\def\grsep{\,\big|\hspace{-3.5pt}\big|\hspace{-3.5pt}\big|\,}

\newcommand{\entails}{\mathbin{|\kern -.47em\sim}}
\newcommand{\notentails}{\mathbin{|\kern -.47em\sim\kern-.9em{/}}}

\def\Lang{\mathcal{L}}

\def\Head{\mathit{Hd}}

\newcommand{\setm}[2]{\{\,#1\mid#2\,\}}

\newcommand{\pf}[1]{\mathit{pf}(#1)}

\def\qed{\hfill $\Box$}
\newcommand\Lb[1]{\mathit{Lb}(#1)}
\newcommand\Hd[1]{\mathit{Head}(#1)}
\newcommand\Bd[1]{\mathit{Body}(#1)}
\def\JM{\mathit{JM}}
\def\SPM{\mathit{SPM}}
\def\CSM{\mathit{CSM}}
\newcommand\forked[1]{#1^{\mid}}
\def\SSM{\mathit{SSM}}

\newenvironment{proofof}[1]{\noindent {\bf Proof of #1.}\\[5pt]}{\qed \vspace{10pt}}
\newtheorem{corollary}{Corollary}

\newtheorem{theorem}{Theorem}
\newtheorem{proposition}[theorem]{Proposition}%
\newtheorem{example}{Example}%
\newtheorem{definition}{Definition}%

\def\JM{\mathit{JM}}
\def\at{\ensuremath{{\cal AT}}}
\newcommand{\eqdef}{%
  \mathrel{\vbox{\offinterlineskip\ialign{%
    \hfil##\hfil\cr%
    $\scriptscriptstyle\mathrm{def}$\cr%
    \noalign{\kern1pt}%
    $=$\cr%
    \noalign{\kern-0.1pt}%
}}}}

\newcommand{\forget}[2]{\mathit{forget}(#1,#2)}

\newcommand{\tuple}[1]{\langle \, #1 \, \rangle}

\begin{document}
\lefttitle{F. Aguado, P. Cabalar, B. Mu\~niz, G. P\'erez and C. Vidal}

\jnlPage{1}{8}
\jnlDoiYr{2025}
\doival{10.1017/xxxxx}

\title[Comparing Non-minimal Semantics for Disjunction in Answer Set Programming]{Comparing Non-minimal Semantics for Disjunction in Answer Set Programming}

\begin{authgrp}
\author{\gn{Felicidad Aguado}$^1$~\orcidlink{0000-0002-4334-9267}}
\affiliation{\email{felicidad.aguado@udc.es}}
\author{\gn{Pedro Cabalar}$^1$~\orcidlink{0000-0001-7440-0953}}
\affiliation{\email{cabalar@udc.es}}
\author{\gn{Brais Mu{\~n}iz}$^1$~\orcidlink{0000-0002-9817-6666}}
\affiliation{\email{brais.mcastro@udc.es}}
\author{\gn{Gilberto P{\'e}rez}$^1$~\orcidlink{0000-0001-6269-6101}}
\affiliation{\email{gperez@udc.es}}
\author{\gn{Concepci{\'o}n Vidal}$^1$~\orcidlink{0000-0002-5561-6406}}
\affiliation{\email{concepcion.vidalm@udc.es}\\[5pt]
$^1$ - University of A Coru\~na, Spain}
\end{authgrp}

\history{\sub{xx xx xxxx;} \rev{xx xx xxxx;} \acc{xx xx xxxx}}

\maketitle

\begin{abstract}
In this paper, we compare four different semantics for disjunction in Answer Set Programming that, unlike stable models, do not adhere to the principle of model minimality.
Two of these approaches, Cabalar and Mu\~niz' \emph{Justified Models} and Doherty and Szalas' \emph{Strongly Supported Models}, directly provide an alternative non-minimal semantics for disjunction.
The other two, Aguado et al's \emph{Forks} and Shen and Eiter's \emph{Determining Inference} (DI) semantics, actually introduce a new disjunction connective, but are compared here as if they constituted new semantics for the standard disjunction operator.
We are able to prove that three of these approaches (Forks, Justified Models and a reasonable relaxation of the DI semantics) actually coincide, constituting a common single approach under different definitions.
Moreover, this common semantics always provides a superset of the stable models of a program (in fact, modulo any context) and is strictly stronger than the fourth approach (Strongly Supported Models), that actually treats disjunctions as in classical logic.
\end{abstract}


\section{Introduction}\label{sec:intro}

\emph{Answer Set Programming} (ASP)~\cite{MT99,Nie99} constitutes nowadays a successful paradigm for practical Knowledge Representation and problem solving.
Great part of this success is due to the rich expressiveness of the ASP language and its declarative semantics, based on the concept of \emph{stable models} in Logic Programming (LP) proposed by Gelfond and Lifschitz~\citeyear{GL88}.
Stable models were originally defined for normal logic programs, but later generalised to accommodate multiple syntactic extensions.
One of the oldest of such extensions is the use of disjunction in the rule heads~\cite{GL91}.
Informally speaking, we may say that the extension of stable models to disjunctive logic programs is based on an \emph{extrapolation of model minimality}.
To explain this claim, let us first recall their original definition for the non-disjunctive case.
To define a stable model $I$ of a program $P$, we first obtain the so-called \emph{program reduct} $P^I$, a program that corresponds to replacing each negative literal in $P$ by its truth value according to $I$.
Program $P^I$ amounts to a set of definite Horn clauses and the semantics for these programs was well-established since the origins of LP.
A (consistent) definite program always has a least model~\cite{vEK76} that further coincides with the the least fixpoint of the \emph{immediate consequences} operator $T_P$, a derivation function that informally corresponds to an exhaustive application of Modus Ponens on the program rules.
A model $I$ of $P$ is \emph{stable} if it coincides with the least model of $P^I$ or, equivalently, the least fixpoint of $T_{P^I}$.
Now, once we introduce disjunction in the rule heads of $P$, the reduct $P^I$ need not be a definite program any more.
As a result, there is no guarantee of a least model (we may have several minimal ones) whereas operator $T_{P^I}$ is not defined, since the application of Modus Ponens may not result in the derivation of atoms\footnote{It is still possible to derive sets of disjunctions of atoms~\cite{LoboMR92}, or sets of minimal interpretations~\cite{FernandezM95} but these options were less explored in the ASP literature.}.
Therefore, two choices are available: (i) requiring $I$ to be \emph{one of the minimal models} of $P^I$; or (ii) modifying the way in which atoms in $P^I$ can be derived, with some alternative to $T_{P^I}$.
As a simple example, consider the disjunctive program $P_{\eqref{f:ab_ac}}$ consisting of rules:
\begin{eqnarray}
a \vee b \hspace{80pt} a \vee c \label{f:ab_ac}
\end{eqnarray}
$P_{\eqref{f:ab_ac}}$ has five classical models $\{a\}$, $\{b,c\}$, $\{a,b\}$, $\{a,c\}$ and $\{a,b,c\}$ but only the first two are minimal.
On the other hand, even though it is a positive program, the application of $T_{P_{\eqref{f:ab_ac}}}$ is undefined and the way to extend it for disjunctive heads is unclear.
Stable models for disjunctive programs~\cite{GL91} adopt criterion (i) based on ``minimality'' -- it is surely the most natural option, but also introduces some drawbacks.
First, we no longer have an associated derivation method like the immediate consequences operator used before.
Second, the complexity of existence of stable model jumps one level in the polynomial hierarchy, from {\sc NP}-complete~\cite{MarekT91} for normal programs to $\Sigma_2^P$-complete for disjunctive programs~\cite{EiterG95}.
%
%

Alternative (ii) has also been explored in the literature in various ways, leading to different disjunctive LP semantics that do not adhere to minimality.
Without trying to be exhaustive, we study here four alternatives that, despite coming from different perspectives, show stunning resemblances.
These four approaches are (by chronological order) the \emph{strongly supported models} by~\citeN{DohertyS15}, the so-called \emph{fork} operators by~\citeN{AguadoCF+19}, the \emph{determining inference (DI) semantics} by~\citeN{ShenE19} the same year, and the \emph{justified models} by~\citeN{CabalarM24}.
In the paper, we prove that the last three cases actually coincide (with a slight relaxation of the DI-semantics), whereas strongly supported models constitute a strictly weaker semantics.
%

The rest of the paper is organised as follows.
The background section contains a description of the approach based on forks which we will take as a reference for most of the correspondence proofs.
It also contains a pair of new results (Section~\ref{sec:or_by_fork}) about replacing disjunctions by forks.
%
%
Section~\ref{sec:JM} describes justified models and proceeds then to prove that the stable models of a fork-based disjunctive program coincide with the justified models of a disjunctive logic program.
In Section~\ref{sec:DI}, we recall the DI-semantics and then prove that, under certain reasonable relaxations of this approach, it also coincides with the semantics of forks.
The next section covers the case of strongly supported models, proving in this case that it constitutes a strictly weaker semantics with respect to the other three approaches, equivalent among them.
Finally, Section~\ref{sec:conclusions} concludes the paper.
The appendices contain the proofs and some previous additional definitions.

\section{Background: overview of Forks}\label{sec:forks}

In this section, we revisit the basic definitions for the fork operators and their denotational semantics.
This semantics is based in its turn on \emph{Equilibrium Logic}~\cite{Pea96} and its monotonic basis, the logic of \emph{Here-and-There} (\HT)~\cite{Hey30}, which is introduced in the first place.
Then, we recall the definition of forks and some previous results that will be used later on for the proofs of correspondence with the other approaches.
Finally, we conclude the section providing a new theorem (Th.~\ref{th:vee_entails_fork}) to be used later, that proves that the replacement of a disjunction by a fork in any arbitrary disjunctive logic program always produces a superset of stable models.

\subsection{Here-and-There and Equilibrium Logic}
Let $\at$ be a finite set of atoms called the \emph{alphabet} or \emph{vocabulary}. A \emph{(propositional) formula} $\varphi$ is defined using the grammar:
\[
\varphi \ ::= \ \ \bot \ \ \grsep \ \ p \ \ \grsep \ \ \varphi \wedge \varphi \ \ \grsep \ \ \varphi \vee \varphi \ \ \grsep \ \ \varphi \to \varphi
\]
where $p$ is an atom $p \in \at$. 
%
%
We use Greek letters $\varphi, \psi, \gamma$ and their variants to stand for formulas. 
We also define the derived operators
\mbox{$(\psi \leftarrow \varphi) \eqdef (\varphi \to \psi)$},
\mbox{$\neg \varphi \eqdef (\varphi \to \bot)$} and
\mbox{$\top \eqdef \neg \bot$}.
%
Given a formula~$\varphi$, by~${\at(\varphi) \subseteq \at}$ we denote the set of atoms occurring in $\varphi$.
%
%
A \emph{theory} $\Gamma$ is a finite\footnote{In this paper, we exclusively focus on finite theories since some of the semantics are not defined for the infinite case.
We leave for future work studying which semantic relations are preserved in that case.
} set of formulas that can be also understood as their conjunction.
When a theory consists of a single formula $\Gamma=\{\varphi\}$ we will frequently omit the braces.
%
An \emph{extended disjunctive rule} $r$ is an implication of the form:
\begin{multline}
\ p_{1} \vee \dots \vee p_{m} \leftarrow p_{m+1}\wedge \dots \wedge p_n \wedge \neg p_{n+1} \wedge \dots \wedge \neg p_h \wedge \neg \neg p_{h+1} \wedge \dots \wedge \neg \neg p_k \label{f:disrule}
\end{multline}
where all $p_i$ above are atoms in $\at$ and $0 \leq m \leq n \leq h \leq k$.
The disjunction in the consequent is called the \emph{head} of $r$ and denoted as $\Hd{r}$, whereas the conjunction in the antecedent receives the name of \emph{body} of $r$ and is denoted by $\Bd{r}$. 
We define the sets of atoms $h(r)\eqdef \{p_{1},\dots,p_m\}$, $b^{+}(r)\eqdef \{p_{m+1},\dots,p_n\}$, $b^{-}(r)\eqdef \{p_{n+1},\dots,p_h\}$, $b^{--}(r)\eqdef\{p_{h+1},\dots,p_k\}$ and $b(r)\eqdef b^{+}(r) \cup b^{-}(r) \cup b^{--}(r)$.
We say that $r$ is an \emph{extended normal rule} if $|h(r)| \leq 1$.
We drop the adjective ``extended'' when the rule does not have double negation.
That is, when $k=h$ we simply talk about a \emph{disjunctive rule} and further call it \emph{normal rule}, if it  satisfies $|h(r)| \leq 1$.
%
%
%
%
%
An empty head $h(r)=\emptyset$ represents falsum $\bot$ and, when this happens, the rule is called a \emph{constraint}.
An empty body $b(r)=\emptyset$ is assumed to represent $\top$ and, when this happens, we usually omit $\leftarrow \top$ simply writing the rule head.
A rule with $b(r)=\emptyset$ and $|h(r)|=1$ is called a \emph{fact}.
\color{brown}
%
\color{black}
A program $P$ is a set of rules and, when the program is finite, we will also understand it as their conjunction.  
We say that program $P$ belongs to a syntactic category if all its rules belong to that category. 

A \emph{classical interpretation} $T$ is a set of atoms ${T \subseteq \at}$. 
By ${T \models \varphi}$ we mean the usual classical satisfaction of a formula $\varphi$. 
Moreover, we write $M(\varphi)$ to stand for the set of  classical models of $\varphi$.
An HT\nobreakdash-\emph{interpretation} is a pair $\tuple{H,T}$ (respectively called ``here'' and ``there'') of sets of atoms ${H \subseteq T \subseteq \at}$; it is said to be \emph{total} when~${H=T}$.
Intuitively, an atom $p$ is considered \emph{false}, when $p \not\in T$, or \emph{true} when $p \in T$, but the latter has two cases: it may be \emph{certainly true} when $p \in H$ or just \emph{assumed true} when ${p \in T \setminus H}$.
An interpretation $\tuple{H,T}$ \emph{satisfies} a formula~$\varphi$, written $\tuple{H,T} \models \varphi$, when the following recursive rules hold:
\begin{itemize}[ topsep=2pt]
\item $\tuple{H,T} \not\models \bot$ 
\item $\tuple{H,T} \models p$ iff $p \in H$ 
\item $\tuple{H,T} \models \varphi \wedge \psi$ iff $\tuple{H,T} \models \varphi$ and $\tuple{H,T} \models \psi$
\item $\tuple{H,T} \models \varphi \vee \psi$ iff $\tuple{H,T} \models \varphi$ or $\tuple{H,T} \models \psi$
\item $\tuple{H,T} \models \varphi \to \psi$ iff both (i) $T \models \varphi \to \psi$, and
\\
\hspace*{132pt}(ii) $\tuple{H,T} \not\models \varphi$ or $\tuple{H,T} \models \psi$
\end{itemize}
An \HT-interpretation $\tuple{H,T}$ is a \emph{model} of a theory $\Gamma$ if $\tuple{H,T} \models \varphi$ for all $\varphi \in \Gamma$.
%
Two formulas (or theories) $\varphi$ and $\psi$ are \HT-equivalent, written $\varphi \equiv \psi$, if they have the same \HT-models.

A total interpretation $\tuple{T,T}$ is an \emph{equilibrium model} of a formula $\varphi$ iff $\tuple{T,T} \models \varphi$ and there is no $H\subset T$ such that $\tuple{H,T} \models \varphi$.
If so, we say that $T$ is a \emph{stable model} of $\varphi$ and we write $\SM{\varphi}$ to stand for the set of stable models of $\varphi$.

\subsection{Forks}\label{subsec:forks}

A \emph{fork} $F$ is defined by the following grammar:
\[
F \ \  ::= \ \  \bot \ \ \grsep \ \ p \ \ \grsep \ \ (F \mid F) \ \ \grsep \ \ F \wedge F \ \ \grsep \ \ \varphi \vee \varphi 
	\ \ \grsep \ \ \varphi \to F
\]
where $\varphi$ is a propositional formula over $\at$ and $p \in \at$ is an atom.
%
It can be proved, by structural induction, that any propositional formula $\varphi$ is a fork.
Note that a fork is not allowed as an argument of a disjunction nor as the antecedent of an implication.
%
%
The intuition of this new connective `$\mid$' is that the stable models of a fork such as $(\varphi_1 \mid \dots \mid \varphi_n)$ -- in fact, all forks are reducible to this form -- will be the \emph{union} of stable models of each $\varphi_i$.
The formal semantics of forks is based on the idea of \emph{denotations} (sets of models) we define next in several steps.

Given a set of atoms $T\subseteq \at$, a $T$-\emph{support}  
\mbox{$\cH \subseteq 2^{T}$} is a set of subsets of $T$ so that, if $\cH \neq \emptyset$, then $T \in \cH$.
Given a propositional formula $\varphi$, the set of sets of atoms $\{H \subseteq T \, \mid \, \tuple{H,T} \models \varphi\}$ forms a $T$-support we denote as~${\den{\varphi}^T}$. 
For readability sake, we directly write a $T$-support as a sequence of sets between square braces: for instance, some possible supports for $T=\{a,b\}$ are $[\{a,b\} \ \{a\}]$, $[\{a,b\}\ \{b\}\ \emptyset]$ or the empty support~$[ \ ]$.
Given two $T$\nobreakdash-supports, $\cH$ and $\cH'$, we define the order relation $\cH \preceq \cH'$ iff either $\cH=[ \ ]$ or $[ \ ] \neq \cH' \subseteq \cH$, read as $\cH$ is ``less supported'' than $\cH'$.
Intuitively, this means that $\cH'$ is closer to make $T$ a stable model than $\cH$.
Given a $T$-support $\cH$, we define its complementary support $\overline{\cH}$ as:
\begin{eqnarray*}
\overline{\cH} \eqdef \left\{
\begin{array}{c@{\ \ }l}
[\ ] & \text{if } \cH=2^T\\
{[\ T\ ]} \cup \{H \subseteq T \mid H \notin \cH\} & \text{otherwise.}
\end{array}
\right.
\end{eqnarray*}

The \emph{ideal} of $\cH$ is defined as $\ideal{\cH} = \{\cH' \mid \cH' \preceq \cH \} \setminus \{ \ [\ ] \ \}$.
Note that, the empty support $[\ ]$ is not included in the ideal, so $\ideal{[ \ ]}=\emptyset$.
If $\Delta$ is any set of supports, we use its $\preceq$-closure:
$$\idealset{\Delta} \eqdef \bigcup_{\cH \in \Delta} \ideal{\cH} = \bigcup_{\cH \in \Delta} \setm{  \cH' \preceq \cH }{\cH' \neq [ \ ]}.$$
We define a \emph{$T$-view} $\Delta$ as any $\preceq$-closed set of $T$\nobreakdash-supports, i.e., $\idealset{\Delta} = \Delta$.
Given a $T$-view $\Delta$, we write 
$\cH \inh \Delta$
iff $\cH \in \Delta$ or both $\cH = [\ ]$ and $\Delta = \emptyset$.

\begin{definition}[{\bf $T$-denotation}]\label{def:fden}
Let $\at$ be a propositional signature and $T \subseteq \at$ a set of atoms.
The $T$-\emph{denotation} of a fork or a propositional formula $F$, written $\fden{F}^T$, is a $T$-view recursively defined as follows:
$$
\begin{array}{ccl}
\fden{\bot}^T &\eqdef& \emptyset
\\[5pt]
\fden{p}^T &\eqdef& \ideal{\den{p}^T} \quad\text{for any atom } p
\\[5pt]
\fden{F \wedge G}^T &\eqdef&
	\idealset{\setm{\cH \cap \cH'}{\cH \in \fden{F}^T \text{ and } \cH' \in \fden{G}^T}}
\\[5pt]
\fden{\varphi \vee \psi}^T &\eqdef&
\idealset{\setm{\cH \cup \cH'}{\cH \inh \fden{\varphi}^T \text{ and } \cH' \inh \fden{\psi}^T}}
\\[5pt]
\fden{\varphi \to F}^T &\eqdef&
	\left\{
	\begin{array}{cl}
	\{ 2^T \} &\text{if } \den{\varphi}^T = [ \ ]
	\\
	\idealset{\setm{\overline{\den{ \varphi}^T } \cup \cH}{ \cH \in \fden{F}^T}} &\text{otherwise}
	\end{array}
	\right.
\\[9pt]
\fden{F \mid G}^T &\eqdef& \fden{F}^T \cup \fden{G}^T 
\end{array}
$$
\end{definition}
\noindent where $F$, $G$ denote forks or propositional formulas.

We say that $T$ is a \emph{stable model} of a fork $F$ when $\fden{F}^T=\ideal{[\, T\, ]}$ or, equivalently, when $[\, T \, ] \in \fden{F}^T$.
%
%
The set $\SM{F}$ collects all the stable models of $F$.

\begin{definition}[Strong Entailment/Equivalence of forks]\label{def:entailment_equivalence_forks}
We say that fork $F$ {\em strongly entails} fork $G$, in symbols $F \entails G$, if $\SM{F \wedge L} \subseteq \SM{G \wedge L}$, for any fork $L$.
We further say that $F$ and $G$ {\em are strongly equivalent}, in symbols $F \cong G$ if both $F \entails G$ and $G \entails F$, that is, $\SM{F \wedge L} =\SM{G \wedge L}$, for any fork $L$.
 \end{definition}
 
Interestingly, \citeN{AguadoCF+19} (Prop.~11) proved that $F \entails G$ is equivalent to $\fden{F}^T \subseteq \fden{G}^T$, for every set of atoms $T \subseteq \at$ and, thus, $F \cong G$ amounts to $\fden{F}^T = \fden{G}^T$, for every $T$.
Other properties proved by \citeN{AguadoCF+19} we will use below are:
\begin{eqnarray}
(F \mid G) \mid L & \cong & F \mid (G \mid L) \label{f:assoc}\\
(F \mid G) \wedge L & \cong & (F \wedge L) \mid (G \wedge L) \label{f:distrib}\\
\SM{F \mid G} & = & \SM{F}\cup \SM{G} \label{f:SMunion}
\end{eqnarray}

\begin{example}\label{ex:forkabc}
Consider the fork:
\begin{eqnarray}
(a \mid b) \wedge (a \mid c) \label{f:forkabac}
\end{eqnarray}
We can apply distributivity \eqref{f:distrib} and associativity \eqref{f:assoc} to conclude that \eqref{f:forkabac} is actually strongly equivalent to:
\begin{eqnarray*}
a \wedge a \mid a \wedge c \mid b \wedge a \mid b \wedge c
\end{eqnarray*}
which is a fork built with 4 propositional formulas.
By \eqref{f:SMunion}, the stable models of this fork are the union of stable models of these 4 formulas, namely, $\{a\}$, $\{a,c\}$, $\{a,b\}$ and $\{b,c\}$.\qed
\end{example}

We conclude this section introducing the polynomial reduction of any fork~$F$ into a propositional formula~$\pf{F}$ by~\citeN{polynomial} that may help for a better understanding of the behaviour of forks, and is used in the proof of Theorem~\ref{th:justified} later on.
For simplicity, we constrained here $\pf{F}$ to the case in which $F$ has the form $\forked{P}$ for some extended disjunctive program $P$, using less definitions and getting  $\pf{F}$ in the form of a disjunctive logic program.
\begin{definition}\label{def:pf}
Let $P$ be some extended disjunctive logic program.
For each $r \in P$ we define $\pf{\forked{r}}$ as: $\pf{\forked{r}}\eqdef r$ if $r$ is an extended normal rule, i.e. $|h(r)|\leq 1$; otherwise, given $h(r)=\{p_1,\dots,p_m\}$:
\[
\pf{\forked{r}} \eqdef 
(x_1 \vee \dots \vee x_m \leftarrow \Bd{r} )\  \wedge \  \bigwedge_{i=1}^m (p_i \leftarrow x_i)
\]
for a set of fresh propositional atoms $x_1, \dots, x_m$.
\qed
\end{definition}
\noindent We also define $\pf{\forked{P}}\eqdef \bigwedge_{r \in P} \pf{\forked{r}}$.
E.g., $\pf{\forked{P}_\eqref{f:ab_ac}}$ is the conjunction of:
\begin{eqnarray*}
x_1 \vee x_2 
\hspace{30pt} 
a \leftarrow x_1 
\hspace{30pt} 
b \leftarrow x_2 
\hspace{30pt} 
y_1 \vee y_2 
\hspace{30pt} 
a \leftarrow y_1
\hspace{30pt} 
c \leftarrow y_2
\end{eqnarray*}

\begin{theorem}
[From Main Theorem~\cite{polynomial}] \label{th:pf}
Let $P$ be an extended logic program.
$\forked{P}$ and $\pf{\forked{P}}$ are strongly equivalent, modulo alphabet $\at(P)$.\qed
\end{theorem}
\subsection{Replacing disjunctions by forks}\label{sec:or_by_fork}

As expected, the definition of stable models for forks is a proper extension of stable models for propositional theories (or if preferred, equilibrium models~\cite{Pea96}) and so, in its turn, it also applies to the more restricted syntax of logic programs with disjunction~\cite{GL91}.
This means that disjunction `$\vee$' in logic programs respects the principle of minimality.
For instance, under this definition we still have the same two stable models for program $P_{\eqref{f:ab_ac}}$, namely, $\SM{P_{\eqref{f:ab_ac}}}=\{\{a\},\{b,c\}\}$.
However, minimality is lost if we replace `$\vee$' by  `$\mid$', as illustrated next.

For any disjunctive rule $r$, let us denote by $\forked{r}$ the fork obtained by substituting in $h(r)$ the operator `$\vee $' by `$\mid$' and let $\forked{P}\eqdef \bigwedge_{r \in P} \forked{r}$ for any program $P$ as expected.
For instance, the fork $\forked{P_{\eqref{f:ab_ac}}}$ would correspond to \eqref{f:forkabac} in Example~\ref{ex:forkabc} whose stable models were $\{a\}$, $\{b,c\}$, $\{a,b\}$ and $\{a,c\}$ -- the last two are not minimal whereas the first two  coincide with $\SM{P_{\eqref{f:ab_ac}}}$.
The main result in this section proves that the replacement of regular disjunctions by forks in any rule $r$ always produces a superset of stable models, even if that rule is included in a larger arbitrary context.
Namely, we have the strong entailment relation $r \entails \forked{r}$.

\begin{theorem}\label{th:vee_entails_fork}
Let $\varphi$ and $\alpha_1, \ldots, \alpha_n$ be propositional formulas with $n\geq 1$.
Then:
$$
\varphi \rightarrow (\alpha_1 \vee \cdots \vee \alpha_n) \; \entails \;  \varphi \rightarrow (\alpha_1 \mid \cdots \mid \alpha_n)  \hspace{50pt}~\Box
$$
\end{theorem}
Since strong entailment allows us to proceed rule by rule, we  conclude:
\begin{corollary}{\label{cor:disjunction_versus_fork}}
Let $P$ be any extended disjunctive logic program, then $P \entails \forked{P}$.\qed
\end{corollary}

As a result, a disjunctive program that has no stable models may restore coherence (existence of stable model) if we replace disjunctions by forks. 
Take the following example (adapted from Ex.~1 by~\citeN{ShenE19}, ).
\begin{example}\label{ex:2}
Consider the program $P_{\eqref{f:SE19}}$ consisting of the three rules:
\begin{eqnarray}
a \vee b \hspace{60pt}
a \hspace{60pt} 
b \leftarrow \neg b \label{f:SE19}
\end{eqnarray}
Disjunction $a\vee b$ is redundant and can be removed, because it is an HT-consequence of $a$.
But once $a \vee b$ disappears, it is clear that $b \leftarrow \neg b$ prevents obtaining any stable model.
Yet, if we change the disjunction in $a \vee b$ by a fork, we can restore coherence. 
The fork $\forked{P_{\eqref{f:SE19}}}$ corresponds to: 
\[
\begin{array}{cl@{\hspace{10pt}}l}
& (a \mid b) \wedge a \wedge (\neg b \to b)\\
\cong & (a \mid b) \wedge a \wedge \neg \neg b & \text{\HT-equivalence}\\
\cong & (a \wedge a \wedge \neg\neg b) \mid (b \wedge a \wedge \neg \neg b) & \text{by distributivity } \eqref{f:distrib}\\
\cong & (a \wedge \neg\neg b) \mid (a \wedge b) & \text{\HT-equivalence}
\end{array}
\]
and then $\SM{P_{\eqref{f:SE19}}}=\SM{a \wedge \neg \neg b} \cup \SM{a \wedge b} = \emptyset \cup \{\{a,b\}\} = \{\{a,b\}\}$, so $\forked{P}_{\eqref{f:SE19}}$ has a unique stable model $\{a,b\}$.
\end{example}
%


\section{Justified Models}\label{sec:JM}
%
We proceed now to compare the forks semantics with  \emph{justified models}~\cite{CabalarM24}.
This approach was originally introduced to provide a definition of \emph{explanations} for the stable models of a logic program.
Such explanations have the form of graphs built with rule labels and reflect the derivation of atoms in the model.
A classical model of a logic program is said to be \emph{justified} if it admits at least one of these explanation graphs.
In the case of normal logic programs, justified and stable models coincide, but  \citeN{CabalarM24} observed that, when the program is disjunctive, it may have more justified models than stable models.
In other words, although every stable model of a disjunctive program admits an explanation, we may have classical models of the program that admit an explanation but are not stable, breaking the principle of minimality in many cases.
In this way, justified models provide a weaker semantics for disjunctive programs that, as we will see, actually coincides with the behaviour of fork-based disjunction.
Let us start recalling some basic definitions, examples and results by \citeN{CabalarM24}.

 \begin{definition}[Labelled logic program]
A \emph{labelled rule} $r$ is an expression of the form $\ell:\eqref{f:disrule}$ where $\eqref{f:disrule}$ is any extended disjunctive rule and $\ell$ is the rule \emph{label}, we will also denote as $\Lb{r}=\ell$.
A \emph{labelled logic program} $P$ is a set of labelled rules that has no repeated labels, that is, for any pair of different rules $r,r' \in P$, $\Lb{r} \neq \Lb{r'}$. 
\end{definition} 

 If $r$ is a labelled rule, we keep the definitions of the formulas $\Bd{r}$ and $\Hd{r}$ and sets of atoms $h(r)$, $b(r)$, $b^+(r)$, $b^- (r)$ and $b^{--}(r)$ as before, that is, ignoring the additional label.
Similarly, if $P$ is a labelled logic program, $\forked{P}$ denotes the fork that results from removing the labels and, as before, replacing disjunctions $\vee$ by $\mid$.
A set of atoms $I$ is a classical model of a labelled rule $r$ iff $I \models \Bd{r} \to \Hd{r}$ in classical logic.
Given a labelled logic program $P$, by $\Lb{P}$ we denote the set of labels of the program $\Lb{P} \eqdef \{\Lb{r}\mid r \in P\}$.
Note that no label is repeated, but $P$ can contain two rules $r,r'$ with the same body and head and different labels $\Lb{r} \neq \Lb{r'}$.
For instance, we could have two repeated facts with different labels $\ell_1:p$ and $\ell_2:p$ possibly representing two different and simultaneously applicable sources of information.

\begin{definition}[Support Graph/Explanation]\label{def:exp}
Let $P$ be a labelled program and $I$ a classical model of $P$. 
A \emph{support graph} $G$ of $I$ under $P$ is a labelled directed graph $G=\tuple{I,E,\lambda}$ where the vertices are the atoms in $I$, the (directed) edges $E \subseteq I \times I$ connect pairs of atoms, and $\lambda: I \to \Lb{P}$ is an injective function that assigns a label for every atom $p \in I$ so that: if $r \in P$ is the rule with $\Lb{r}=\lambda(p)$ then $p \in h(r)$, $I \models \Bd{r}$ and $b^{+}(r) = \{ q \mid (q,p) \in E \}$.
A support graph $G$ is said to be an \emph{explanation} if it additionally satisfies that $G$ is acyclic.\qed
\end{definition}
The fact that $\lambda$ is injective guarantees that there are no repeated labels in the graph.
Additionally, the definition tells us that if an atom $p$ is labelled with $\lambda(p)=\ell$ then $\ell$ must be the label of some rule $r$ where (1) $p$ occurs in the head, (2) the body of the rule is satisfied by $I$ and (3) the incoming edges for $p$ are formed from the atoms in the positive body of $r$.
Since an explanation $G=\tuple{I,E,\lambda}$ for a model $I$ is uniquely determined by its atom labelling $\lambda$, we can abbreviate $G$ as a set of pairs $p \mapsto \lambda(p)$ for $p \in I$.

\begin{definition}[Supported/Justified model]
\label{def:justified}
Let $I$ be classical model of a labelled program $P$, $I \models P$.
Then, $I$ is said to be a (\emph{graph-based}) \emph{supported model} of $P$ if there exists some support graph of $I$ under $P$, and is further said to be a \emph{justified model} of $P$ if there exists some explanation (i.e. acyclic support graph) of $I$ under $P$.
Sets $\SPM(P)$ and $\JM(P)$ respectively stand for the (graph-based) supported and justified models of $P$.\qed
\end{definition}

We can also define $\SPM(P)$ and $\JM(P)$ for any non-labelled program $P$ by assuming we previously label each rule in $P$ with a unique arbitrary identifier.
Note that different labels produce different explanations, but the definition of justified/supported model is not affected by that.

%
%
%
%
\begin{theorem}[Th.~1 and Th.~2 from \citeN{CabalarM24}]\label{th:justified_Brais}
If $P$ is a labelled disjunctive program, then: $ \SM{P}\subseteq \JM(P)$.
Moreover, if $P$ contains no disjunction, then $ \SM{P}=\JM(P).$ \qed
\end{theorem}

However, if we allow disjunction, we may have justified models that \emph{are not} stable models, as illustrated below.
\begin{example}
Let $P_\eqref{f:ab_ac2}$ be the following labelled version of $P_\eqref{f:ab_ac}$:
 \begin{eqnarray}
\ell_1: a \vee b \hspace{70pt}
\ell_2: a \vee c \label{f:ab_ac2}
\end{eqnarray}
The classical models of $P_\eqref{f:ab_ac2}$ are $\{a\}, \{a,b\}, \{a,c\}, \{b,c\}, \{a,b,c\}$. 
The last one, $\{a,b,c\}$, is not justified, since we would need three different labels and we only have two rules.
Each model $\{a,c\}$, $\{a,b\}$, $\{b,c\}$ has a unique explanation corresponding to the atom labellings $\{a \mapsto \ell_1, c \mapsto \ell_2\}$, $\{b \mapsto \ell_1, a \mapsto \ell_2\}$ and $\{b \mapsto \ell_1, c \mapsto \ell_2\}$, respectively.
On the other hand, model $\{a\}$ has two possible explanations, corresponding to $\{ a \mapsto \ell_1\}$ and $\{a \mapsto \ell_2\}$.
To sum up, we get four justified models, $\{a,c\}$, $\{a,b\}$, $\{b,c\}$ and $\{a\}$ but only two of them are stable, $\{a\}$ and $\{b,c\}$.
\qed
\end{example}
In other words, the justified models of $P_\eqref{f:ab_ac2}$ coincide with the stable models of its fork version $\forked{P_\eqref{f:ab_ac2}}=\forked{P_\eqref{f:ab_ac}}=\eqref{f:forkabac}$ seen before.
This is in fact, a general property that constitutes the main result of this section.

\begin{theorem}\label{th:justified}
$\JM(P) = \SM{\forked{P}}$ for any labelled disjunctive logic program $P$. \qed
\end{theorem}

Supported models $\SPM(P)$ correspond to the case in which we also accept cyclic explanation graphs.
Obviously, $\JM(P) \subseteq \SPM(P)$, because all acyclic explanations are still acceptable for $\SPM(P)$.
\citeN{CabalarM24} also proved that $\SPM(P)$ generalise the standard notion of supported models -- i.e., models of Clark's completion~\cite{Cla78} -- to the disjunctive case.
For instance, the program $P_\eqref{f:ploop}$ consisting of the rule:
\begin{eqnarray}
\ell_1: p \leftarrow p \label{f:ploop}
\end{eqnarray}
has two supported models, $I=\emptyset$ (which is also stable and justified) and $I=\{p\}$ with a cyclic support graph where node $p$ is connected to itself.
As a remark, notice that the definition of our ``graph-based'' \emph{supported models}~\cite{CabalarM23} does not correspond to the (also called) supported models obtained from the \emph{program completion} defined by~\citeN{AD16} for disjunctive programs.
The latter, we denote $AD(P)$, impose a stronger condition: a rule $r$ supports an atom $p \in \Head(r)$ with respect to interpretation $I$ not only if $I \models \Bd{r}$ but also $I \not\models q$ for all $q \in \Head(r) \setminus{p}$.
To illustrate the difference, take program $P_\eqref{f:ab_ac2}$: as it has no cyclic dependencies, graph-based supported and justified models coincide, that is, $\SPM(P_\eqref{f:ab_ac2})=\JM(P_\eqref{f:ab_ac2})=\{\{a\},\{a,b\},\{a,c\},\{b,c\}\}$ we saw before.
However, $AD(P_\eqref{f:ab_ac2})=\{\{a\},\{b,c\}\}$ that, in this case, coincide with the stable models of the program.
Model $\{a,b\}$ is supported (under Def,~\ref{def:justified}) because $a$ is justified by rule $\ell_1$ and $b$ by rule $\ell_2$.
However, under Alviano and Dodaro's definition, rule $\ell_1$ is not a valid support for $a$ since we would further need $b$ (the other atom in the disjunction $\ell_1$) to be false.
The situation for model $\{a,c\}$ is analogous.
It is not hard to see that $\SM{P} \subsetneq AD(P) \subsetneq \SPM(P)$ (the first inclusion proved by~\citeN{AD16}), so clearly, $AD(P)$ is more interesting for computation purposes when our goal is approximating $\SM{P}$.
However, $\SPM(P)$ provides a more liberal generalisation of the definition of supported model from normal logic programs: as in that case, $I$ is a supported model of $P$ if, for every atom $p \in I$, there exists some rule $r$ with $p$ ``in the head" and $I\models \Bd{r}$. 
This definition has also a closer relation to $\JM(P)$ and explanation generation or to the DI semantics (as we see in Theorem~\ref{th:supported} in the next section).

\section{Determining Inference}\label{sec:DI}

The third approach we consider, \emph{Determining Inference} (DI)~\cite{ShenE19}, also introduces a new disjunction operator in rule heads, with the same syntax as forks `$|$'.
Besides, first-order formulas are allowed to play the role of atoms, and so, the syntax accepts regular disjunction `$\vee$' too.
However, in this paper (for the sake of comparison) we  describe the DI-semantics directly on the syntax of extended disjunctive rules of the form~\eqref{f:disrule} seen before\footnote{To be precise, \citeN{ShenE19} treat double negation classically, whereas here, we take the liberty to keep doubly negated atoms in the body and interpret them as in Equilibrium Logic.}, using `$\vee$' to play the role of the DI disjunctive operator.

The DI-semantics understands disjunction as a non-deterministic choice and is based on the definition of a \emph{head selection function}.
This function will tell us, beforehand, which head atom will be chosen if we have to apply a rule for derivation.
We introduce next a slight generalisation of that definition.

\begin{definition}[Open/Closed Head Selection Function]\label{def:sel}
Let $P$ be an extended disjunctive logic program and $I \subseteq \at$ an interpretation.
A \emph{head selection function} $sel$ for $I$ and some $r \in P$ is a formula:
$$
sel(\Hd{r},I) \eqdef \left \{ 
\begin{array}{ll}
\bot & \mbox{ if } h(r) \cap I = \emptyset\\
p_i & \mbox{ otherwise, for some } p_i \in h(r) \cap I
\end{array}
\right.
$$
We say that $sel$ is \emph{closed} if $sel(\Hd{r},I)=sel(\Hd{r'},I)$ for any pair of rules $r,r'$ with the same head atoms $h(r)=h(r')$.
If this restriction does not apply, we just say that $sel$ is \emph{open}. \qed
\end{definition}
The original definition by~\citeN{ShenE19} (Def.~4) corresponds to what we call here \emph{closed} selection function and forces the same choice when two rule heads are formed by the same set of atoms.

The \emph{reduct} of a program $P$ with respect to some interpretation $I$ and selection function $sel$ is defined as the logic program $
P_{sel}^{I} \eqdef \{\  sel(\Hd{r},I) \leftarrow \Bd{r} \mid  I \models \Bd{r} \ \}
$.
Note that $P^I_{sel}$ is an extended normal logic program (possibly containing constraints) where we replaced each disjunction by the atom determined by the selection function $sel$.
%
\begin{definition}[Candidate stable model]\label{def:1}
A classical model $I$ of a an extended disjunctive logic program $P$ is said to be a \emph{candidate stable model}\footnote{Or \emph{candidate answer set} in the original terminology~\cite{ShenE19}.} if there exists a selection function $sel$ such that
\mbox{$I \in \SM{P^I_{sel}}$}.
We further say that $I$ is \emph{closed} if $sel$ is closed.
By $\CSM(P)$, we denote the set of candidate stable models of $P$.
\qed
\end{definition}
To understand the difference between closed and open selection functions, take the following program $P_\eqref{f:closed}$:
\begin{eqnarray}
p \hspace{60pt} \bot \leftarrow c \hspace{60pt} a \vee b \hspace{60pt} b \vee a \leftarrow p \label{f:closed}
\end{eqnarray}
The set $\CSM(P_\eqref{f:closed})$ consists of $\{p, a\}$, $\{p, b\}$ and $\{p,a,b\}$, but only the first two models are closed, since they make the same choice in both disjunctions $a \vee b$ and $b \vee a$ that have the same atoms.
Note that this condition is rather syntax-dependent: if we replace $b \vee a \leftarrow p$ by the rule $b \vee a \vee c \leftarrow p$, then open candidate stable models are not affected ($c$ must be false due to constraint $\bot \leftarrow c$) but $\{p,a,b\}$ becomes now a closed candidate stable model, since the sets of atoms in $a \vee b$ and $b \vee a \vee c$ are different.

A \emph{DI-stable model} $I$ of a program $P$ is a model that is minimal among the closed candidate stable models (Def.~7 by \citeN{ShenE19}).
Thus, DI-semantics actually imposes an additional minimality condition.
However, if we focus on the previous step, $\CSM(P)$, we can prove that they coincide with $\SM{\forked{P}}$ and, by Theorem~\ref{th:justified}, with $\JM(P)$ too.


\begin{theorem}\label{th:candidate}
 $\CSM(P)=\SM{\forked{P}}$ for any extended disjunctive logic program $P$.\qed
\end{theorem}

We conclude this section by proving that the decision problem $\CSM(P)\neq \emptyset$ is \textsc{NP}-complete, recalling the following complexity result proved by~\citeN{ShenE19} 

\begin{proposition}[From Table~1 by~\citeN{ShenE19}]\label{prop:NP}
Deciding the existence of a DI-stable model for a disjunctive program, under the well-justified semantics~\cite{SHEN14}, is an NP-complete problem.
\end{proposition}

\begin{theorem}\label{th:NP}
Given an extended disjunctive logic program $P$, deciding $\CSM(P)\neq \emptyset$ is an NP-complete problem.
\end{theorem}

As one last result in this section, we provide an alternative characterisation of the supported models from Def.~\ref{def:justified} using DI semantics.
For normal logic programs, $I$ is a supported model of $P$ if, for every atom $p \in I$, there exists some rule $r$ with $p$ in the head and $I\models \Bd{r}$.
Alternatively, supported models can also be captured as the fixpoints of the immediate consequences~\cite{vEK76} operator $T_P\eqdef \{p \mid (p \leftarrow B) \in P, I \models B\}$, namely, $I$ is a supported model of $P$ iff $I=T_P(I)$.
We can extend this relation for disjunctive logic programs as follows.

\begin{theorem}\label{th:supported}
Let $P$ be a labelled program and $I$ a classical model of $P$. The following assertions are equivalent:
\begin{enumerate}
    \item $I \in \SPM(P)$
    \item $T_{P_{sel}^I}(I)=I$ for some head selection function $sel$.
\end{enumerate}

\end{theorem}

\section{Strongly supported models}\label{sec:SSM}
For our last comparison, we consider \emph{strongly supported models} by~\citeN{DohertyS15}:

\begin{definition}[Strongly Supported Models\footnote{We use Def.4 by~\citeN{DohertyKS16} but adjusting $H_0$ as in Def.~11 by~\citeN{DohertyS15}.}]
\label{def:ssm}
A model $T$ of an extended disjunctive logic program $P$  is a \emph{strongly supported model} of $P$ if there exists a sequence of interpretations $H_0 \subseteq H_1 \subseteq \dots \subseteq H_n=T$ such that
\begin{enumerate}
\item For $i=0$: $H_0 \cap h(r) \neq \emptyset$ for all $r \in P$ with $b(r)=\emptyset$.\\
For $i \geq 1$: $H_i \cap h(r) \neq \emptyset$ for all $r \in P$ with
\footnote{The original definition is not given in terms of \HT-satisfaction, but it uses a definition involving pairs of sets of atoms that is completely equivalent, for the syntactic fragment of logic programs.
}
$\tuple{H_{i-1},T} \models \Bd{r}$.
\item For each $i\geq 0$: $H_i$ only contains atoms obtained by applying point 1, that is, if $p \in H_i$ then $p \in h(r)$ for some rule $r$ mentioned in point 1.
\end{enumerate}
We denote the set of strongly supported models of $P$ as $\SSM(P)$.\qed
\end{definition}

\citeN{DohertyS15} (Th.~1) proved that the stable models of $P$, $\SM{P}$, coincide with the minimal elements of $\SSM(P)$ and, furthermore, $\SM{P}=\SSM(P)$ when $P$ has no disjunction.
However, in general, the $\SSM$ semantics makes disjunction to behave classically.
For instance, from Def.~\ref{def:ssm} above, we can easily observe that, if $P$ is a set of disjunctions of atoms, then  $\SSM(P)=M(P)$.
As a result, since $\eqref{f:ab_ac}$ is a pair of disjunctions, $\SSM(P_\eqref{f:ab_ac})=M(P_\eqref{f:ab_ac})$ i.e., the five classical models $\{a\}$, $\{a,b\}$, $\{a,c\}$, $\{b,c\}$ and $\{a,b,c\}$ mentioned before.
Note that $\CSM$ did not accept $\{a,b,c\}$, pointing our that it is a stronger semantics, as corroborated next:

\begin{theorem}\label{th:ssm}
$\CSM(P) \subseteq \SSM(P)$ for any extended disjunctive logic program $P$.\qed
\end{theorem}

To conclude this section, we observe that, despite their name similarity, supported $\SPM(P)$ and strongly supported models $\SSM(P)$ are unrelated.
To prove $\SPM(P) \not\subseteq \SSM(P)$, just take the program $P_\eqref{f:ploop}$ with no disjunctions, so that $\SSM(P)=\SM{P}=\{\emptyset\}$.
However, $\{p\} \in \SPM(P)$ as we discussed before.
To prove $\SSM(P) \not\subseteq \SPM(P)$ we already saw that $\{a,b,c\} \in \SSM(P_\eqref{f:ab_ac}) \setminus \JM(P_\eqref{f:ab_ac})$.
But, since $P_\eqref{f:ab_ac}$ has no implications, the support graphs contain no edges, so that acyclicity is irrelevant meaning $\JM(P_\eqref{f:ab_ac})=\SPM(P_\eqref{f:ab_ac})$.

\section{Conclusions}~\label{sec:conclusions}

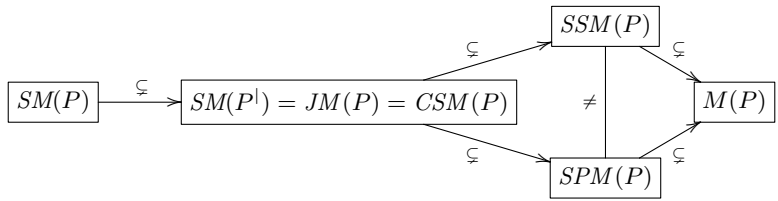
\begin{figure}
\centering
\[
\xymatrix@-4mm {
& & & 
*+[F]{\SSM(P)} 
\ar@{-}[dd]_{\neq}
\ar[rd]^{\subsetneq} 
\\
*+[F]{\SM{P}} \ar[rr]^(0.3){\subsetneq} 
& &
*+[F]{\SM{\forked{P}}=\JM(P)=\CSM(P)}
\ar[ru]^{\subsetneq}
\ar[rd]_{\subsetneq} 
&
&
*+[F]{M(P)}
\\
& & & 
*+[F]{\SPM(P)}
\ar[ru]_{\subsetneq} 
}
\]
\caption{Inclusion relations among several semantics for disjunctive logic programs.}
\label{fig:relations}
\end{figure}

We have studied four different semantics for any disjunctive logic program $P$ in Answer Set Programming (ASP) that, unlike the standard stable models $\SM{P}$ do not adhere to the principle of model minimality.
These four approaches are: \emph{forks}~\cite{AguadoCF+19} here denoted as $\SM{\forked{P}}$; \emph{justified models}~\cite{CabalarM24} $\JM(P)$; (a relaxed version of) \emph{determining inference} ~\cite{ShenE19} we denoted $\CSM(P)$; and \emph{strongly supported models} $\SSM(P)$~\cite{DohertyS15}.
The summary of our results is shown in Figure~\ref{fig:relations}, where $M(P)$ represents the classical models of $P$ and $\SPM(P)$ an extension of supported models for the disjunctive case~\cite{CabalarM24}.
Interestingly, the three semantics $\SM{\forked{P}}$, $\JM(P)$ and $\CSM(P)$ coincide, although their definitions come from rather different approaches, showing that they may capture a significant way to understand disjunction in ASP, removing minimality and keeping the computational complexity of existence of stable model as an \textsc{NP}-complete problem.

For future work, we plan to study other alternatives.
For instance, one reviewer suggested replacing disjunctions by choice rules~\cite{SNS02} so that each disjunctive rule of the form
$p_1 \vee \dots \vee p_m \leftarrow \mathit{Body}$ becomes the choice rule $1 \{ p_1, \dots , p_m \} \leftarrow \mathit{Body}$
and the rest of rules are left untouched.
The behaviour of this replacement produces similar results to $\SSM(P)$ and we plan to study a formal (dis)proof of this coincidence for future work.

\noindent {\bf Acknowledgements}. We wish to thank the anonymous reviewers for their useful comments that have helped to improve the paper and, especially, for refuting an incorrect proof of a result included in a previous version of the document.
This research was partially funded by the Spanish Ministry of Science, Innovation and Universities, MICIU/AEI/
10.13039/501100011033, grant PID2023-148531NB-I00, Spain.

\bibliographystyle{acmtrans}
\bibliography{refs}

\newpage
\appendix
\section{Additional definitions}

The proof of Theorem~\ref{th:justified} requires some additional definitions by~\citeN{AguadoCF+19} and~\citeN{CabalarM23} that we include in this section.
We start introducing some concepts for the denotational semantics that allow us to work with a sub-alphabet $V \subseteq \at$.
We define $\SMR{\varphi}{V} \, \eqdef \, \{\, T \cap V \mid T \in \SM{\varphi}\,\}$ as the projection of $\SM{\varphi}$ onto some vocabulary $V$.
For any $T$-support $\cH$ and set of atoms $V$, we write $\cH_V$ to stand for $\{H \cap V \mid H \in \cH\}$.
We say that a $T$-support $\cH$ is $V$-\emph{feasible} iff there is no ${H \subset T}$ in $\cH$ satisfying that $H \cap V = T \cap V$.
The name comes from the fact that, if this condition does not hold for some $\cH=\den{\varphi}^T$ with $H \subset T$, then $T$ cannot be stable for any formula ${\varphi \wedge \psi}$ with $\psi \in {\mathcal L}_V$ because $H$ and $T$ are indistinguishable for any formula $\psi \in {\mathcal L}_V$.

If $F$ is a fork and $T\subseteq V\subseteq \at$, we can define the $T$-view:
$$
\fden{F}^{T}_{V} \eqdef \idealset{\setm{\cH_V}{ \cH \in \fden{F}^Z \mbox{ s.t. }Z \cap V= T \mbox{ and } \cH \mbox{ is } V\mbox{-feasible} }}.
$$

Given a fork $F$, the set $\SMR{F}{V}$ denotes the projection of 
$\SM{F}$ on the set of atoms $V$, that is 
$\SMR{F}{V} \eqdef \{T \cap V \mid T \in \SM{F}\}$.

\begin{definition}[Projective Strong Entailment/Equivalence of forks]\label{def:projective_entailment_equivalence_forks}
Let $F$ and $G$ be forks and $V \subseteq \at$ a set of atoms.
We say that $F$ {\em strongly $V$-entails} $G$, in symbols $F \entails_{V} G$, if for any fork $L$ in $\Lang_V$,  $\SMR{F \wedge L}{V} \subseteq \SMR{G \wedge L}{V}$. We further say that $F$ and $G$ {\em are strongly $V$-equivalent}, in symbols $F \cong_{V} G$ if both $F \entails_{V} G$ and $G \entails_{V} F$, that is, $\SMR{F \wedge L}{V} =\SMR{G \wedge L}{V}$, for any fork $L$ in $\Lang_V$.
 \end{definition}

Note that, when $\at(F) \cup \at(G) \subseteq V$, $F \entails_V G$ and $F \cong_V G$ respectively amount to the non-projective definitions $F \entails G$ and $F \cong G$ we introduced in Definition~\ref{def:entailment_equivalence_forks}.

Now, the idea of projecting out some atoms of the alphabet can also be applied to explanation graphs, as introduced by~\citeN{CabalarM23} (Def.~5) under the name of \emph{node forgetting}. 

\begin{definition}[Node forgetting]\label{def:forgetting}
Let $G=\tuple{I,E,\lambda}$ be an explanation and let $A \subseteq \at$ be a set of atoms to be removed.
Then, the \emph{node forgetting} operation on $G$ produces the graph $\forget{G}{A}\eqdef \tuple{I\setminus A,\ E',\ \lambda}$ where $E'$ contains all edges $(p_0,p_n)$ such that there exists a path $(p_0,p_1), (p_1,p_2), \dots, (p_{n-1},p_n)$ in $E$ with $n\geq 0$, $\{p_0,p_n\} \cap A = \emptyset$ and $\{p_1,\dots,p_{n-1}\} \subseteq A$.
\end{definition}

Note that $E'$ contains all original edges $(p,q) \in E$ for non-removed atoms $\{p,q\} \cap A = \emptyset$, since they correspond to the case where $n=1$.
%
It is clear that, if $G$ is acyclic, then $\forget{G}{A}$ is also acyclic. The only new edges in $\forget{G}{A}$ are $(p_0,p_n)$ such that there exists a path $(p_0,p_1)$, $(p_1,p_2)$, \dots, $(p_{n-1},p_n)$ in $E$. This implies that, if there exist a cycle in $\forget{G}{A}$, then we will also have a cycle in $G$.

\section{Proofs}

\begin{proofof}{Theorem~\ref{th:vee_entails_fork}}
Given that $F \entails G$ amounts to $\fden{F}^T \subseteq \fden{G}^T$ for any $T$, we have to prove: 
$$
\fden{\varphi \rightarrow (\alpha_1 \vee \cdots \vee \alpha_n)}^T \; \subseteq \; \fden{ \varphi \rightarrow (\alpha_1 \mid \cdots \mid \alpha_n)}^T  $$
for any $T$.
Given the denotation of implication (Def.~\ref{def:fden}) this amounts to proving $\fden{ \alpha_1 \vee \cdots \vee \alpha_n}^T \, \subseteq \, \fden{ \alpha_1 \mid \cdots \mid \alpha_n}^T$.
Unfolding these two expressions:
\begin{eqnarray*}
\fden{ \alpha_1 \vee \cdots \vee \alpha_n}^T   & = &   \ideal{\den{\alpha_1 \vee \cdots \vee \alpha_n}^T} = \ideal{\big(\den{\alpha_1}^T \cup  \ldots \cup  \den{\alpha_n}^T\big)}\\
\fden{ \alpha_1 \mid \cdots \mid \alpha_n}^T  & = &    \fden{\alpha_1}^T \cup \ldots \cup \fden{\alpha_n}^T = \ideal{{\den{(\alpha_1)}^T}} \cup \ldots \cup \ideal{{\den{(\alpha_n)}^T}} 
\end{eqnarray*}
we essentially have to prove:
$$
\ideal{\big(\den{\alpha_1}^T \cup  \ldots \cup  \den{\alpha_n}^T\big)} \subseteq \ideal{{\den{(\alpha_1)}^T}} \cup \ldots \cup \ideal{{\den{(\alpha_n)}^T}}
$$
We can distinguish two cases:
\begin{enumerate}
\item Suppose $T \models \alpha_i$ for some $i$, $1\leq i \leq n$.
Then, $T \in \den{\alpha_i}^T \neq [\ ]$ whereas, obviously, $\den{\alpha_i}^T  \subseteq \den{\alpha_1}^T \cup  \ldots \cup  \den{\alpha_n}^T$ because $1 \leq i \leq n$.
As a result, $\big(\den{\alpha_1}^T \cup  \ldots \cup  \den{\alpha_n}^T \big ) \,  \preceq \, \den{\alpha_i}^T$.
This implies:
$$
\ideal{\big(\den{\alpha_1}^T \cup  \ldots \cup  \den{\alpha_n}^T\big)} \, \subseteq \,  \ideal{\den{\alpha_i}^T} \, \subseteq \,  \ideal{{\den{(\alpha_1)}^T}} \cup \ldots \cup \ideal{{\den{(\alpha_n)}^T}}.
$$

\item Suppose $T \not\models \alpha_i$ for all $i$, $1\leq i \leq n$.
This means $\den{\alpha_i}^T=[\ ]$ for all $i$.
But, then $\ideal{\big(\den{\alpha_1}^T \cup  \ldots \cup  \den{\alpha_n}^T\big)}=\ideal [\ ] = \emptyset \subseteq \ideal{{\den{(\alpha_1)}^T}} \cup \ldots \cup \ideal{{\den{(\alpha_n)}^T}}$ since the empty support $[\ ]$ is never included in the ideal so $\ideal [\ ] = \emptyset$.
\end{enumerate}
\end{proofof}

%

\begin{proofof}{Theorem~\ref{th:justified}}
{\bf From right to left}: $\JM(P)\subseteq \SM{\forked{P}}$. Suppose $I=\{a_1,a_2,\ldots,a_n\} \in \JM(P)$ and so,  $G=\tuple{I,E,\lambda}$ is an explanation of $I$ under $P$. For any $1 \leq i \leq n$, if $\lambda(a_i)=Lb(r_i)$, we know that $a_i \in h(r_i)$ and $I \models \Bd{r_i}$. We are going to prove that, for any $r \in P$, we can find $\cH_r \in \fden{\forked{r}}^{I}$ such that:
$$
[\, I \, ]= \bigcap_{r\in P}\cH_r
$$
which would mean that $I \in \SM{\forked{P}}$ by applying Definition~\ref{def:fden}. Notice that $I \models P$, so, for any $r \in P$ such that $I \models \Bd{r}$, there exists $a_r \in h(r)$ such that $a_r \in I$. Let us define: 
$$
    \cH_r:= \left\{ \begin{array}{ll}
\overline{\den{\Bd{r}}^I} \cup \den{a_r}^I & \mbox{ if } I \models \Bd{r} \\
2^I & \mbox{ otherwise }
\end{array}
\right.
    $$
Notice that $\cH_r \in \fden{\forked{r}}^{I}$ and suppose that $J \in \cH_r$, for any $r \in P$.
If $J \neq I$, we get $J \subset I$ (since $\cH_r$ is an $I$-support) and, so there exists $a_0 \in I \setminus J$. 
Since $a_0 \in I \in \JM(P)$, there is a label $\lambda(a_0)=\Lb{r_0}$ for some rule $r_0$ with $a_0 \in h(r_0)$ and $I \models \Bd{r_0}$.
The latter implies $\cH_{r_0}=\overline{\den{\Bd{r_0}}^I} \cup \den{a_0}^I$.
Since $J \in \cH_{r_0}$ and $a_0 \not\in J$ we conclude $\tuple{J,I} \not\models \Bd{r_0}$.
Since $I \models \Bd{r_0}$ then we can find $a_1 \in b^{+}(r_0) \setminus J$. 
As a result, there is an edge $(a_1,a_0)\in E$.
We can repeat the same reasoning to get $a_1$ from $a_0$ and continue like this finding $a_1, a_2$, $a_3$, \dots, such that $a_i \in b^{+}(r_{i-1}) \setminus J$ and getting the edges $(a_i,a_{i-1}) \in E$ for $i \geq 1$. 
Since $I$ is finite, we know that, for some $j,k$, $0 \leq j < k \leq |I|$ such that $a_{j}=a_{k}$.
Notice that we have found a cycle in $G$ given by the path $a_{k},a_{k-1}, \dots, a_{j} $ which yields a contradiction.

\noindent {\bf From right to left}: $\SM{\forked{P}} \subseteq \JM(P)$.
Take $I \in \SM{\forked{P} }$.
To prove $I \in \JM(P)$, since $P$ is not a labelled program, we start  assigning an arbitrary and unique label $\Lb{r}:=\ell_r$ for every rule $r \in P$.
We must then find some explanation $G=\tuple{I,E,\lambda}$ of $I$ under (labelled) $P$.
We will obtain this explanation through another labelled program that results from a particular labelling of $\pf{\forked{P}}$, that is, the translation of $\forked{P}$ into a extended disjunctive logic program.
According to Def.~\ref{def:pf}, $\pf{\forked{P}}$ is the result of replacing every rule $r \in P$ with $|h(r)|>1$ by the rules:
\begin{eqnarray}
x_1 \vee \dots \vee x_m & \leftarrow & \Bd{r} \label{f:pf1}\\
p_i \leftarrow x_i \label{f:pf2}
\end{eqnarray}
for each $r \in P$ with $h(r)=\{p_1,\dots,p_m\}$ and for all $i=1,\dots,m$, being $x_i$ new fresh atoms for each rule $r$.
We propose the following labelling for program $\pf{\forked{P}}$:
\begin{itemize}
\item If $|h(r)|=1$ then $\Lb{r}=\ell_r$
\item Otherwise, $\Lb{\eqref{f:pf1}}:=\ell^0_r$, $\Lb{\eqref{f:pf2}}:=\ell^i_r$ for all $i=1,\dots,m$, where all these labels $\ell^0_r, \ell^i_r$ are fresh for each rule $r$.
\end{itemize}
Let us call $Aux:=\at(\pf{\forked{P}}) \setminus \at(P)$, that is, the collection of all fresh variables $x_i$ introduced in $\pf{\forked{P}}$.
Since, by Theorem~\ref{th:pf}, ${\pf{\forked{P}} \cong_{\at(\forked{P})} \forked{P}},$ we know there exists $\hat{I}\in \SM{\pf{\forked{P}}}$ such that $\hat{I} \cap \at(\forked{P})=I.$
Notice that $\hat{I} \in \JM(\pf{\forked{P} })$ by Theorem~\ref{th:justified_Brais}, so there exists a labelled graph $\hat{G}=\tuple{\hat{I},\hat{E}, \hat{\lambda}}$ which is an explanation of $\hat{I}$ under $\pf{\forked{P} }$. 
Consider the graph $\forget{\hat{G}}{Aux}=\tuple{(\hat{I} \setminus Aux),E}=\tuple{I,E}$. 
In order to complete an explanation of $I$ under $P$, we need a map $\lambda: I \to Lb(P)$ satisfying the requirements of Definition~\ref{def:exp}. 
Take any $p \in I \subseteq \hat{I}$ and its label $\hat{\lambda}(p)$ in $\hat{G}$.
We can distinguish two cases:

\begin{itemize}
\item $\hat{\lambda}(p)$ is the label of some rule $r \in \pf{\forked{P}}$ that was not modified from $P$, that is $r \in P$, because $h(r)=\{p\}$.
This means $\hat{\lambda}(p)=\ell_r$ and in this case we keep the same label $\lambda(p):=\ell_r$.
\item Otherwise, $p$ came from the head of some rule $\eqref{f:pf2}$ and so $\hat{\lambda}(p)$ has the form $\ell^i_r$ for some $i=1,\dots,|h(r)|$.
Then, we take $\lambda(p):=\ell_r$.
\end{itemize}
We will proceed to show that the labelled graph formed by $G=\tuple{I,E, \lambda}$ is an explanation of $I$ under $P$.
First, notice that, $\forget{\hat{G}}{Aux}$ is acyclic because $\hat{G}$ was acyclic and node forgetting maintains acyclicity.
Second, we observe next that $\lambda$ is injective.
By contradiction, suppose we had two different atoms $p,q \in I$ and $\lambda(p)=\lambda(q)$.
We have two cases: if $\lambda(p)=\hat{\lambda}(p)$ and $\lambda(q)=\hat{\lambda}(q)$ then, as $\hat{\lambda}$ is injective, we conclude $p=q$ reaching a contradiction.
Otherwise, without loss of generality, suppose $\lambda(p)\neq \hat{\lambda}(p)$ (the proof for $q$ is analogous).
Then $\hat{\lambda}(p)=\ell^i_r$ and $\lambda(p)=\ell_r=\lambda(q)$.
Therefore, $q$ is also labelled in $\lambda$ with the same rule $r$ and $|h(r)|>1$.
By construction of $\lambda$, this implies $\hat{\lambda}(q)=\ell^j_r$ for some other $j \neq i$, since $p$ and $q$ are different atoms.
Now, consider the rules of the form $\eqref{f:pf2}$ for the labels we have obtained $\ell_r^i$ for $p$ and $\ell_r^j$ for $q$.
But those atoms only occur in the head of one rule of form $\eqref{f:pf1}$ that has the label $\ell^0_r$.
So, we have concluded $\ell^0_r=\hat{\lambda}(x_i)=\hat{\lambda}(x_j)$ and, since $\hat{\lambda}$ is injective $x_i=x_j$ so $i=j$ reaching a contradiction.

Finally, we have to prove that $\lambda$ respects the rest of conditions of Definition~\ref{def:exp}, namely, that for each atom $p \in I$, given the rule $r \in P$ such that $\lambda(p)=\Lb{r}=\ell_r$ then: (i) $I \models \Bd{r}$; and (ii) there is an edge $(q,p) \in E$ for all $q \in b^+(r)$.

For proving (i), we will show that $\hat{I} \models \Bd{r}$, something that implies $I \models \Bd{r}$ because $\hat{I} \setminus Aux = I$ and $b(r)\cap Aux = \emptyset$.
We have again two cases: when $h(r)=\{p\}$ we just have $\hat{\lambda}(p)=\lambda(p)=\ell_r$ and the rule $r \in P$ also appears unchanged in $\pf{\forked{P}}$ with the same label.
Since $\hat{G}$ is an explanation for $\hat{I}$, we conclude $\hat{I}\models \Bd{r}$.
Otherwise $|h(r)|>1$ and  $\hat{\lambda}(p)=\ell^i_r$ for some $i=1,\dots,|h(r)|$.
But then, we have an edge $(x_i,p) \in \hat{E}$ and $\hat{\lambda}(x_i)=\ell^0_r$ for a rule of the form \eqref{f:pf2} for the same $r \in P$.
Note that $\hat{I}$ must satisfy the body of \eqref{f:pf2} that actually coincides with $\Bd{r}$, so $\hat{I} \models \Bd{r}$ again.

For proving (ii),
Finally take $q \in b^+(r)$ for some rule $r$ such that $\lambda(p)=\Lb{r}=\ell_r$ with $p \in I$. If $h(r)=\{p\}$, then $\hat{\lambda}(p)=\ell_r$, so $(q,p) \in \hat{E}$ and also $(q,p) \in E$ by construction of $\forget{\hat{G}}{Aux}$ since $\{q,p\}  \subseteq \at(P)$.
On the other hand, suppose  $\lambda(p)=\Lb{r}=\ell_r$ for $r \in P$ with $|h(r)|>1$. Then $\hat{\lambda}(p)=\ell^i_r$ and we have the labelled rule $\ell^i_r: p \leftarrow x_i$ for some $i=1,\dots,|h(r)|$.
Therefore $x_i \in \hat{I}$ and $(x_i,p) \in \hat{E}$.
Moreover, the only possible label for $x_i$ corresponds to the only rule where it appears $\hat{\lambda}(x_i)=\ell^0_r$.
Since the body of that rule is $\Bd{r}$, we must also have $(q,x_i) \in \hat{E}$ as $q \in b^+(r)$.
Finally, since both $(q,x_i)$ and $(x_i,p)$ belong to $\hat{E}$, by construction of $\forget{\hat{G}}{Aux}$ we conclude $(q,p) \in E$ because $x_i \in Aux$ whereas $p,q \not\in Aux$.
\end{proofof}


\begin{proofof}{Theorem~\ref{th:candidate}}
\noindent {\bf From left to right} $\CSM(P) \subseteq \SM{\forked{P}}$: suppose that $I \in \CSM(P)$, that is, there exists some function $sel$ for which $I\in \SM{P_{sel}^{I}}$. 
We have to prove $I \in \SM{\forked{P}}$ or, if preferred, $[I] \in \fden{\forked{P}}^I$.
Since $\forked{P} = \bigwedge_{r \in P} \forked{r}$, this amounts to proving that we have some combination of $I$-supports $\cH_r \in \fden{\forked{r}}^I$ per each $r \in P$ such that $\bigcap_{r \in p} \cH_r = [I]$.
In particular, for each $r \in P$, take the $I$-support:
$$
\cH_r:=\left\{ 
\begin{array}{ll}
   \overline{\den{\Bd{r}}^I} \cup \den{sel(\Hd{r},I)}^I & \mbox{ if } I \models \Bd{r}\\
    2^I & \mbox{ otherwise }
\end{array}
\right.
$$
We see next that, according to Definition~\ref{def:fden}, $\cH_r \in \fden{\forked{r}}^I$.
If $I \models \Bd{r}$, we get $I \models sel(\Hd{r},I)$ because $I\models P_{sel}^{I}$, $\forked{r} \in P_{sel}^{I}$ and $\forked{r}=(sel(\Hd{r},I) \leftarrow \Bd{r})$. 
Note that, $sel(\Hd{r},I) \in I$ is some atom in the head of $r$ and so, is one of the branches of the head fork in $\forked{r}$.
On the other hand, if $I \not\models \Bd{r}$, $\cH_r=2^I$ and this is trivially included in the ideal $\fden{\forked{r}}$ as it is the $\preceq$-bottom element.
To prove $[I]=\bigcap_{r \in p} \cH_r$ we proceed by contradiction: suppose we had a different $J \neq I$, such that $J \in \cH_r$ for all $r \in P$.
As $J$ belongs to $I$-supports, we conclude $J \subset I$.
But, since $I$ is a stable model of $P_{sel}^{I}$, there must be some rule $r' \in P_{sel}^{I}$ for which $\tuple{J,I} \not\models P_{sel}^{I}$.
Suppose $r'$ has the form $sel(\Hd{r},I) \leftarrow \Bd{r}$ for one of the rules $r \in P$.
Since $I \models r'$ the only possibility is $\tuple{J,I} \models \Bd{r}$ but $\tuple{J,I} \not\models sel(\Hd{r},I)$.
This implies both $J \in \den{\Bd{r}}^I$ and $J \not\in \den{sel(\Hd{r},I)}^I$ which is in contradiction with $J \in \cH_r$.

\noindent {\bf From right to left} $\SM{\forked{P}} \subseteq \CSM(P)$: take some $I\in \SM{\forked{P}}$.
To prove $I \in \CSM(P)$ we must find some head selection function $sel$ for $I$ such that $I \in \SM{P^I_{sel}}$.
Consider the translation of $\forked{P}$ into a propositional formula $pf(\forked{P})$, in our case, the disjunctive logic program we saw in Def.~\ref{def:pf}.
%
%
Remember that $\pf{\forked{P}}$ is the result of replacing every rule $r \in P$ with $|h(r)|>1$ by the rules \eqref{f:pf1} and \eqref{f:pf2} for each $r \in P$ with $h(r)=\{p_1,\dots,p_m\}$ and for all $i=1,\dots,m$, being $x_i$ new fresh atoms for each rule $r$.
By Theorem~\ref{th:pf}, there exists some stable model $\hat{I}$ of $pf(\forked{P})$ such that $\hat{I} \cap \at = I$.
In fact, it is not hard to see that, for each rule $r$ such that $\hat{I} \models \Bd{r}$, there exists a \emph{unique} auxiliary atom $x_i \in \hat{I}$ from the head of \eqref{f:pf1}.
This is because, if we had two atoms $x_i, x_j \in \hat{I}$ with $i \neq j$ then the HT interpretation $\tuple{\hat{I}\setminus \{x_j\},\hat{I}}$ would still satisfy $pf(\forked{P})$: rule \eqref{f:pf1} is still satisfied because $x_i \in  \hat{I}\setminus \{x_j\}$ whereas $p_j \leftarrow x_j$ is trivially satisfied since $x_j \not\in \hat{I}\setminus \{x_j\}$ whereas $\hat{I} \models (p_j \leftarrow x_j)$.
Note also that $\hat{I} \models \Bd{r}$ iff $I \models \Bd{r}$ because all atoms occurring in $\Bd{r}$ belong to $\at$.
Let us define the following head selection function:
$$
sel(\Hd{r},I) \eqdef \left \{ 
\begin{array}{ll}
\bot & \mbox{if } h(r) \cap I = \emptyset \\
p_i & \mbox{if }  h(r) \cap I \neq \emptyset, I \models \Bd{r}, \\
    & (p_i \leftarrow x_i) \in pf(\forked{P})\ \text{and } x_i \in \hat{I}\\
p_j & \mbox{if }  h(r) \cap I \neq \emptyset, I \not\models \Bd{r}, \\
    & \text{for some } p_j \in  h(r) \cap I
\end{array}
\right.
$$
We can check that this is a valid selection function, since if selects some head atom among those in $h(r) \cap I$, when this set is not empty.
In particular, when $I \models \Bd{r}$ (and so $\hat{I} \models \Bd{r}$) we fix the selected atom to $p_i$ for the corresponding auxiliary atom $x_i$ occurring in $\hat{I}$.
We know $p_i \in I$ because $p_i \in \hat{I}$ and this atom is not auxiliary, $\hat{I} \cap \at = I$.
On the other hand, when $I \not\models \Bd{r}$, the selection function is irrelevant, since it is not used in $P^I_{sel}$.
For proving $I \in \SM{P^I_{sel}}$ we first observe that $I \models P^I_{sel}$ holds by construction of this program because it contains rules of the form $p \leftarrow \Bd{r}$ where $r \in P$, $I \models \Bd{r}$ and $p \in h(r) \cap I$.
By contradiction, suppose we had some $J \subset I$, $\tuple{J,I} \models P^I_{sel}$.
Then, take:
$$
\hat{J}:=J \cup \{x_i \in \hat{I} \mid p_i \in J, (p_i \leftarrow x_i) \in pf(\forked{P})\}
$$
Clearly, $\hat{J} \subset \hat{I}$ and so $\tuple{\hat{J},\hat{I}} \not\models pf(\forked{P})$ because $\hat{I}$ is a stable model of that program.
Then, there must be some rule in $pf(\forked{P})$ not satisfied by $\tuple{\hat{J},\hat{I}}$ and notice that all rules $r \in P$ with $|h(r)|\leq 1$ are left untouched both in $pf(\forked{P})$ and $P^I_{sel}$, whereas they do not contain auxiliary atoms, so $\tuple{J,I} \models r$ iff $\tuple{\hat{J},\hat{I}} \models r$.
In other words, we must have some rule of the form \eqref{f:pf1} or \eqref{f:pf2} in $pf(\forked{P})$ not satisfied by $\tuple{\hat{J},\hat{I}}$.
Suppose $\tuple{\hat{J},\hat{I}} \not\models p_i \leftarrow x_i$ for some of the rules of the form \eqref{f:pf2}.
Then $p_i \not\in \hat{J}$ and $x_i \in \hat{J}$ but this is impossible by construction of $\hat{J}$.
Suppose, instead, that $\tuple{\hat{J},\hat{I}} \not\models x_1 \vee \dots \vee x_m \leftarrow \Bd{r}$ for some of the rules of the form \eqref{f:pf1}.
This implies $x_j \not\in \hat{J}$ for all $j=1,\dots,m$ and $\tuple{\hat{J},\hat{I}} \models \Bd{r}$.
Since $\Bd{r}$ has no auxiliary atoms, the latter implies $\tuple{J,I} \models \Bd{r}$.
As $\tuple{J,I} \models sel(\Hd{r},I) \leftarrow \Bd{r}$ we conclude $sel(\Hd{r},I)$ is some atom $p_k \in h(r) \cap J \subset J'$ for some $k=1,\dots,m$.
But then, by construction of $\hat{J}$ we should have $x_k \in \hat{J}$ for some $k=1,\dots,m$ and we reach a contradiction.  
\end{proofof}

\begin{proofof}{Theorem~\ref{th:NP}}
The result follows from Proposition~\ref{prop:NP} by~\citeN{ShenE19} but making some minor considerations.
The application of Shen and Eiter's result has three main differences with respect to what we try to prove: (i) it refers to DI-stable models, which are minimal among the candidate stable models; (ii) it uses well-justified (WJ) semantics to interpret the program $P^I_{sel}$; (iii) it takes \emph{closed} candidate stable models.
Difference (i) is not important, because there exists some minimal candidate stable model iff there exists some candidate stable model (assuming finite interpretations, so we always have a minimal interpretation among a set of interpretations).
Namely, the problem of existence of a closed candidate stable model for a disjunctive program using WJ semantics is also NP-complete.

Similarly, difference (ii) is not too relevant either:  the specific use of the WJ semantics is important when the program allows arbitrary formulas in the rule bodies, something we do not consider in the current paper.
For normal logic programs, WJ collapses to standard stable models.
Yet, in the current paper, we do allow $P^I_{sel}$ to be an \emph{extended} normal logic program, that is, we allow double negation in the bodies.
However, we can reduce $P$ to an equivalent program without double negation (modulo auxiliary atoms) while keeping the HT semantics.
We define $t_1(P)$ as the result of replacing each doubly negated literal $\neg \neg q$ by $\neg aux$ in all rule bodies and adding the rule $aux \leftarrow \neg q$, where $aux$ is a new fresh atom per each different doubly negated literal.
Program $t_1(P)$ is strongly equivalent to program $P$ modulo auxiliary atoms~\cite{CFCPV17} and is obtained by a linear transformation.
So the existence of a closed candidate stable model for an \emph{extended} disjunctive logic program is also an NP-complete problem.

Finally, difference (iii) can also be easily overcome: given a disjunctive program $P$, we can we can obtain its $CSM(P)$ by computing the closed CSMs of another program $t_2(P)$ defined as follows. 
For each set of rules $r_1, r_2, \dots, r_n$ sharing the same set of head atoms $\Head(r_1)=\Head(r_2)=\dots=\Head(r_n)$ replace each rule head of the form $\Hd{r_i}=p_1 \vee \dots \vee p_m$ by the disjunction $p_1 \vee \dots p_m \vee aux_i$ and add the constraint $\bot \leftarrow aux_i$ to the new program, where $aux_i$ is a new, fresh atom per each one of these rules.
Program $t_2(P)$ has the same candidate stable models as $P$ (auxiliary atoms are always false) but never repeats the sets of head atoms in disjunctions, so its closed CSMs are just the $CSM(t_2(P))$.
Note that transformation $t_2(P)$ is polynomial (we can apply some ordering algorithm on the rules and rule heads to check coincidence of head atoms).

To sum up, deciding $CSM(P)\neq \emptyset$ can be reduced in polynomial time to deciding the existence of a closed candidate stable model of $t_2(t_1(P))$ and the latter is an NP-complete problem according to Proposition~\ref{prop:NP}.
\end{proofof}

\begin{proofof}{Theorem~\ref{th:ssm}}
Let $T \in \CSM(P)$. Then, there exists some head selection function $sel$ for $P$ and $T$ such that $T$ is a stable model of $P^T_{sel}$. 
To prove $T \in \SSM(P)$, we must find a  sequence of interpretations $H_0 \subseteq H_1 \subseteq \dots \subseteq H_n=T$ for some $n\geq 0$ satisfying the conditions in Definition~\ref{def:ssm}.
Take the following sequence:
\begin{eqnarray*}
H_0 & := & \{sel(\Hd{r},T) \mid r \in P \mbox{ and } b(r)=\emptyset \}; \\[5pt]
H_{i} & := & H_{i-1} \cup \{sel(\Hd{r},T) \mid r \in P, \tuple{H_{i-1}\cap T,T} \models \Bd{r} \},\\
& & \text{for } i>0.
\end{eqnarray*}
We can see that $H_{i-1} \subseteq H_i$ by definition, and so, assuming a finite signature, there exists some $n\geq 0$ such that $H_n=H_{n+1}$.
We prove next $\bot \not\in H_i \subseteq T$ for all $i \geq 0$ by induction.
For the base case, take any $sel(\Hd{r},T) \in H_0$.
Since it is obtained from a rule with empty body $b(r)=\emptyset$, we trivially have $T \models \Bd{r}\ (\equiv \top)$ so  $sel(\Hd{r},T)$ becomes a fact in the reduct $P^T_{sel}$.
Since $T \models P^T_{sel}$, we get $sel(\Hd{r},T) \in T$ and, moreover, $sel(\Hd{r},T)\neq \bot$ must be some head atom from $r$.
Now, for $i>0$, note that $H_i$ consists of the union of $H_{i-1} \subseteq T$ (by induction) and $\{sel(\Hd{r},T) \mid r \in P, \tuple{H_{i-1}\cap T,T} \models \Bd{r} \}$.
Take any $sel(\Hd{r},T)$ in this last set and let $r$ be the rule satisfying $\tuple{H_{i-1}\cap T,T} \models \Bd{r}$.
By the Persistence property in HT, $\tuple{H_{i-1}\cap T,T} \models \Bd{r}$ implies $T \models \Bd{r}$ classically, and so, the rule $sel(\Hd{r},T) \leftarrow \Bd{r}$ is included in $P^T_{sel}$.
Again, as $T \models P^T_{sel}$, and $T \models \Bd{r}$ we conclude $sel(\Hd{r},T) \in T$ and $sel(\Hd{r},T) \neq \bot$.

Now, since all $H_i \subseteq T$, we can actually change the condition $\tuple{H_i \cap T, T} \models \Bd{r}$ simply by $\tuple{H_i, T} \models \Bd{r}$ as in Def.~\ref{def:ssm}:
\begin{eqnarray*}
H_{i} & := & H_{i-1} \cup \{sel(\Hd{r},T) \mid r \in P, \tuple{H_{i-1},T} \models \Bd{r} \} \quad \text{for } i>0
\end{eqnarray*}
Note that these sets so defined $H_0 \subseteq H_1 \subseteq \dots \subseteq H_n$ satisfy now Conditions 1 and 2 in Def.~\ref{def:ssm} by construction.
Condition 1 holds because all atoms in $H_i$ are selected among the heads $h(r)$ of the same corresponding rules $r \in P$ (remember $sel$ never produces $\bot$ in $H_i$), whereas Condition 2 is satisfied because no other atoms are included in $H_i \setminus H_{i-1}$.
However, to obtain $T \in \SSM(P)$, we still remain to prove that $H_n=T$ and  we only know $H_n \subseteq T$.
By contradiction, suppose $H_n \subset T$.
Since $T$ is a stable model of $P^T_{sel}$, we know $\tuple{H_n,T} \not\models P^T_{sel}$.
That is, there is some rule $r \in P$,  $r^T_{sel} \in P^T_{sel}$ of the form $r^T_{sel}=(sel(\Hd{r},T) \leftarrow \Bd{r})$ such that $\tuple{H_n,T} \not\models r^T_{sel}$ and $T \models \Bd{r}$.
Since $T \models P^T_{sel}$, the only possibility is $\tuple{H_n,T}\models \Bd{r}$ but $\tuple{H_n,T} \not\models sel(\Hd{r},T)$.
From $\tuple{H_n,T}\models \Bd{r}$ and the definition of $H_{n+1}$ we conclude $sel(\Hd{r},T) \in H_{n+1}=H_n$, reaching a contradiction.
\end{proofof}

\begin{proofof}{Theorem~\ref{th:supported}}
Suppose that there exists some head selection function $sel$ such that  $T_{P_{sel}^I}(I)=I$. 
Let us define a supported graph $G=\tuple{I,E,\lambda}$ of $I$ under $P$. For any $p \in I$, there exists $\Bd{r} \to sel(\Hd{r},I) \in P_{sel}^I$ such that $I \models \Bd{r}$ and $sel(\Hd{r},I)=p$. 
We define $\lambda(p)=\Lb{r}$. 
If $\lambda(p)=\Lb{r_1}=\lambda(q)=\Lb{r_2}$, then $r_1=r_2$ (different rules have different labels), so $p=sel(\Hd{r_1},I)=sel(\Hd{r_2},I)=q$. 
By definition of $\lambda$, if $\lambda(p)=\Lb{r}$, then $p=sel(\Hd{r},I) \in h(r)$ for some rule $r\in P$ such that $I \models \Bd{r}.$ 
This means that $G$ is a supported graph of $I$ under $P.$

\par On the other hand, suppose that $I \in \SPM(P)$ being $G=\tuple{I,E,\lambda}$ a supported graph of $I$ under $P$.
Let us define the following head selection function:
$$
sel(\Hd{r},I) \eqdef \left \{ 
\begin{array}{ll}
\bot & \mbox{if } h(r) \cap I = \emptyset \\
p & \mbox{if }  h(r) \cap I \neq \emptyset, I \models \Bd{r}, \\
    &  \text{and } \lambda(p)=\Lb{r}\\
q & \mbox{if }  h(r) \cap I \neq \emptyset, I \models \Bd{r}, \\
    & \text{for any } q \in  h(r) \cap I \ \text{ such that } \lambda(q) \neq \Lb{r} \\
    r & \mbox{if }  h(r) \cap I \neq \emptyset, I \not \models \Bd{r}, \\
    & \text{for any } r \in  h(r) \cap I\\
\end{array}
\right.
$$
We have defined $sel(\Hd{r},I)$ in such a way that, whenever $I \models \Bd{r}$ and there is an atom $p \in h(r)  \cap I$ such that $\lambda(p)=\Lb{r}$,  we force $sel(\Hd{r},I)=p$. 
Notice that, there can not be two different atoms $p,q \in h(r)$ with $\lambda(p)=\lambda(q)=\Lb{r}$ because $\lambda$ is injective. 
In case there is no such atom $p \in h(r)$ with $\lambda(p)=\Lb{r}$, we take another atom $q \in h(r) \cap I$ which always exists because $I \models P$.

We always have that $sel(\Hd{r},I) \in I$ when $I \models \Bd{r}$, so $T_{P_{sel}^I}(I) \subseteq I$.
Moreover, if $p \in I$, we know that $\lambda(p)=\Lb{r_0}$ for some $r_0 \in P$ such that $I \models \Bd{r_0}$ and $p \in h(r_0)$. 
The definition of our $sel$ function implies that $sel(\Hd{r_0}),I)=p \in T_{P_{sel}^I}(I)$.
\end{proofof}
\end{document}